\documentclass[letterpaper, 10 pt, conference]{ieeeconf}

\IEEEoverridecommandlockouts                              % This command is only needed if 
\usepackage{cite}
\usepackage{graphics} % for pdf, bitmapped graphics files
\usepackage{epsfig} % for postscript graphics files
\usepackage{amssymb}  % assumes amsmath package installed
\usepackage{amsmath} % assumes amsmath package installed
\usepackage{mathtools}
\usepackage{color}
\usepackage[vietnamese, english]{babel}
\usepackage[T1]{fontenc}
\usepackage{bm}
\usepackage{hyperref}

\usepackage[nolist]{acronym}

\newcommand{\norm}[1]{||#1||} 

\renewcommand{\vec}[1]{\bm{#1}}

\newcommand{\systemname}[0]{multi-robot system}

\newcommand{\PREPRINTYEAR}{2024}
\newcommand{\PUBLISHEDIN}{IEEE/RSJ International Conference on Intelligent Robots and Systems (IROS)}
\newcommand{\DOI}{10.1109/IROS58592.2024.10802469} % you will not get a DOI until the paper is actually published, so update this when you get it and reupload the new preprint to all systems

\newcommand{\reffig}[1]{Fig.~\ref{#1}}

\newcommand{\refsec}[1]{Sec.~\ref{#1}}

\renewcommand{\refeq}[1]{\eqref{#1}}
\newcommand{\worldframe}[0]{\mathcal{W}}
\newcommand{\worldframeaxis}[1]{\mathbf{w}_{#1}}
\newcommand{\bodyframe}[1]{\mathcal{B}_{#1}}
\newcommand{\bodyframeaxis}[2]{\mathbf{b}_{#1,#2}}
\newcommand{\uavforce}[0]{\vec{F}_{u}}
\newcommand{\boatforce}[0]{\vec{F}_{b}}
\newcommand{\vesselparallelframe}[0]{\mathcal{VP}}
\newcommand{\vesselparallelframeaxis}[1]{\mathbf{vp}_{#1}}
\usepackage[binary-units=true]{siunitx}
\usepackage{pgfplots}
\usetikzlibrary{quotes, angles, backgrounds, arrows, automata, shapes, positioning, calc, through, spy, decorations.pathreplacing, decorations.markings, arrows.meta, automata, petri, intersections, pgfplots.fillbetween}

\usepackage{tikz}
\usepackage{pgfplots}
\pgfplotsset{compat=1.14}
\usetikzlibrary{backgrounds,arrows,automata,shapes,positioning,calc,through,spy}
\pgfdeclarelayer{background}
\pgfdeclarelayer{foreground}
\pgfsetlayers{background, main, foreground}
\tikzset{
  state/.style={
    rectangle,
    draw=black, very thick,
    minimum height=1.0em,
    text centered,
  },
  smallstate/.style={
    rectangle,
    draw=black, very thick,
    minimum height=0.2em,
    text centered,
  },
  final_state/.style={
    rectangle,
    rounded corners,
    draw=black, very thick,
    minimum height=2em,
    text centered,
  },
  initial_state/.style={
    rectangle,
    double=white,
    double distance=1pt,
    inner sep=2pt,
    draw=black, very thick,
    minimum height=2em,
    text centered,
  },
  point/.style={
    circle,
    inner sep=0pt,
    minimum size=3pt,
    fill=red
  },
  adder/.style={
    circle,
    inner sep=2pt,
    minimum size=0.3in,
    draw=black, very thick,
    text centered
  },
  state_gray/.style={
    rectangle,
    draw=black, very thick,
    fill=gray!40,
    minimum height=1.0em,
    text centered,
    inner sep=0,
  },
  state_white/.style={
    rectangle,
    draw=black, very thick,
    fill=white,
    minimum height=1.0em,
    text centered,
    text=black,
    inner sep=0,
  },
  state_green/.style={
    rectangle,
    draw=black, very thick,
    fill=green!50,
    minimum height=1.0em,
    text centered,
    text=black,
    inner sep=0,
  },
  state_red/.style={
    rectangle,
    draw=black, very thick,
    fill=red!70,
    minimum height=1.0em,
    text centered,
    text=black,
    inner sep=0,
  },
  state_blue/.style={
    rectangle,
    draw=black, very thick,
    fill=blue!40,
    minimum height=1.0em,
    text centered,
    text=black,
    inner sep=0,
  },
  final_state/.style={
    rectangle,
    rounded corners,
    draw=black, very thick,
    minimum height=2em,
    text centered,
  },
  initial_state/.style={
    rectangle,
    double=white,
    double distance=1pt,
    inner sep=2pt,
    draw=black, very thick,
    minimum height=2em,
    text centered,
  },
  point/.style={
    circle,
    inner sep=0pt,
    minimum size=3pt,
    fill=red
  },
}

\tikzset{new spy style/.style={spy scope={%
  magnification=5,
  size=1.25cm,
  connect spies,
  every spy on node/.style={
    rectangle,
    draw,
  },
  every spy in node/.style={
    draw,
    rectangle,
    fill=white
  }
  }
  }
}
\title{\LARGE \bf
Collaborative Object Manipulation on the Water Surface by a UAV-USV Team Using Tethers  
}

\author{Filip Nov\'{a}k$^{1}$, Tom\'{a}\v{s} B\'{a}\v{c}a, and Martin Saska % <-this % stops a space
    \thanks{Authors are with the Multi-robot Systems Group, Department of Cybernetics, Faculty of Electrical Engineering, Czech Technical University in Prague, Czech Republic, {\tt\small \{filip.novak, tomas.baca, martin.saska\}@fel.cvut.cz}.}
    \thanks{$^{1}$Corresponding author}
    \thanks{This work was funded by the Czech Science Foundation (GA\v{C}R) under research project no. 23-07517S, by the European Union under the project Robotics and advanced industrial production (reg. no. CZ.02.01.01/00/22\_008/0004590), and by CTU grant no SGS23/177/OHK3/3T/13.}
}

\usepackage[placement=top,vshift=-7]{background}
\SetBgScale{1.0}
\SetBgContents{IROS 2024. PREPRINT VERSION - DO NOT DISTRIBUTE. \href{https://doi.org/\DOI}{DOI \DOI}}
\SetBgColor{black}
\SetBgAngle{0}
\SetBgOpacity{1.0}

\begin{document}

% Place this in the document body as the first part to have an extra first page with the copyright. Remove this part to get rid of the extra page.
\thispagestyle{empty}
\onecolumn
{
  \topskip0pt
  \vspace*{\fill}
  \centering
  \LARGE{%
    \copyright{} \PREPRINTYEAR~\PUBLISHEDIN\\\vspace{1cm}
    Personal use of this material is permitted.
    Permission from \PUBLISHEDIN~must be obtained for all other uses, in any current or future media, including reprinting or republishing this material for advertising or promotional purposes, creating new collective works, for resale or redistribution to servers or lists, or reuse of any copyrighted component of this work in other works.\vspace{1cm}\newline
    \large DOI: \href{https://doi.org/\DOI}{\DOI}
    \vspace*{\fill}}
    \vspace*{\fill}
}
\NoBgThispage
\twocolumn          	% Comment out for single-column articles
\BgThispage

\maketitle
\thispagestyle{empty}
\pagestyle{empty}

%%%%%%%%%%%%%%%%%%%%%%%%%%%%%%%%%%%%%%%%%%%%%%%%%%%%%%%%%%%%%%%%%%%%%%%%%%%%%%%%
\begin{abstract}
This paper introduces an innovative methodology for object manipulation on the surface of water through the collaboration of an Unmanned Aerial Vehicle (UAV) and an Unmanned Surface Vehicle (USV) connected to the object by tethers.
We propose a novel mathematical model of a robotic system that combines the UAV, USV, and the tethered floating object.
A novel Model Predictive Control (MPC) framework is designed for using this model to achieve precise control and guidance for this collaborative robotic system.
Extensive simulations in the realistic robotic simulator Gazebo demonstrate the system's readiness for real-world deployment, highlighting its versatility and effectiveness.
Our multi-robot system overcomes the state-of-the-art single-robot approach, exhibiting smaller control errors during the tracking of the floating object's reference.
Additionally, our multi-robot system demonstrates a shorter recovery time from a disturbance compared to the single-robot approach.
\end{abstract}

%%%%%%%%%%%%%%%%%%%%%%%%%%%%%%%%%%%%%%%%%%%%%%%%%%%%%%%%%%%%%%%%%%%%%%%%%%%%%%%%
\section{INTRODUCTION}
The utilization of heterogeneous unmanned vehicles and systems cooperating in maritime operations has garnered growing interest, especially when considering teams of \acp{USV} and \acp{UAV}.
The \ac{UAV}-\ac{USV} teams have already proven its usefulness in various fields, including
search and rescue operations \cite{usv_uav_hurricane_wilma, usv_uav_hurricane_wilma2, xiao2017UAVAssistedUSV, usv_uav_rescue_system}, the search and removal of pollutants from water surfaces \cite{usv_uav_water_pollution, han2021AutomaticMonitoringWater, deng2022AutomaticCollaborativeWater}, hydrologic data collection \cite{usv_uav_measurement}, coastal activities monitoring \cite{wu2023CooperativeUnmannedSurface}, and detection and monitoring of marine fauna \cite{verfuss2019ReviewUnmannedVehicles}.

One of the challenges encountered during marine robotics applications is the manipulation and transportation of objects in water, which is relevant to numerous marine robotics applications, ranging from debris removal \cite{brandao2022SidePullManeuverNovel} to the precise measurement of water environments using sensors \cite{kapetanovic2021HeterogeneousRoboticSystem}.
In scenarios such as waste collection, the ability to tow objects is evident, while in environmental monitoring, sensors require controlled movement along predefined paths or proximity adjustments to specific points of interest.
Additionally, it is crucial to guarantee the protection of transported objects, such as expensive sensors, preventing any potential damage or destruction during maneuvers around obstacles. 
Therefore, the implementation of precise movement control for all system components becomes imperative.
Recognizing that a single type of robot may prove inefficient for such varied tasks, our approach involves the collaboration of a team of heterogeneous robots.

Previous research has highlighted the potential of \acp{UAV} to tow objects \cite{kourani2021BidirectionalManipulationBuoy, kourani2023ThreedimensionalModelingTethered}.
While \acp{UAV} excel in maneuverability and in providing an aerial perspective of the surrounding water surface, they are constrained by lower force and a shorter flight duration.
In a complementary manner, \acp{USV} designed for towing waterborne sensors have demonstrated extended mission duration and increased payload capacity \cite{manley2016UnmannedSurfaceVessels, meng2019StudyMechanicalCharacteristics}.
However, \acp{USV} are constrained by low maneuverability and have a restricted field of view over the surrounding water surface due to their limited height above the surface \cite{xiao2017UAVAssistedUSV}, where obstacles may mutually obstruct each other or remain concealed beneath the surface.
To harness the advantages of both robots, we propose a unified system where an object is tethered to a \ac{USV} and to a \ac{UAV} simultaneously.
% This integration improves precision, maneuverability, power, and situational awareness from an elevated perspective.
% The maneuverability of the \ac{UAV} integrated into a robotic system is useful during dynamic obstacle avoidance.
Moreover, the tether between the \ac{USV} and \ac{UAV} can provide a reliable wired communication link \cite{xu2021ApplicationResearchTethered} and energy transfer from the \ac{USV} to the object (e.g., sensor) and the \ac{UAV}, significantly prolonging the \ac{UAV} mission duration \cite{choi2014TetheredAerialRobots}.
The collaboration of \ac{UAV} and \ac{USV} enhances maneuverability, towing force, precision in controlling the movements of manipulated objects, and provides situational awareness from an elevated perspective.
Our exploration of this tethered system aims to showcase its potential in overcoming the limitations of individual robots.
%, offering a comprehensive solution for diverse marine robotics applications.

\begin{figure}[!t]
    \centering
    \includegraphics[width=\linewidth]{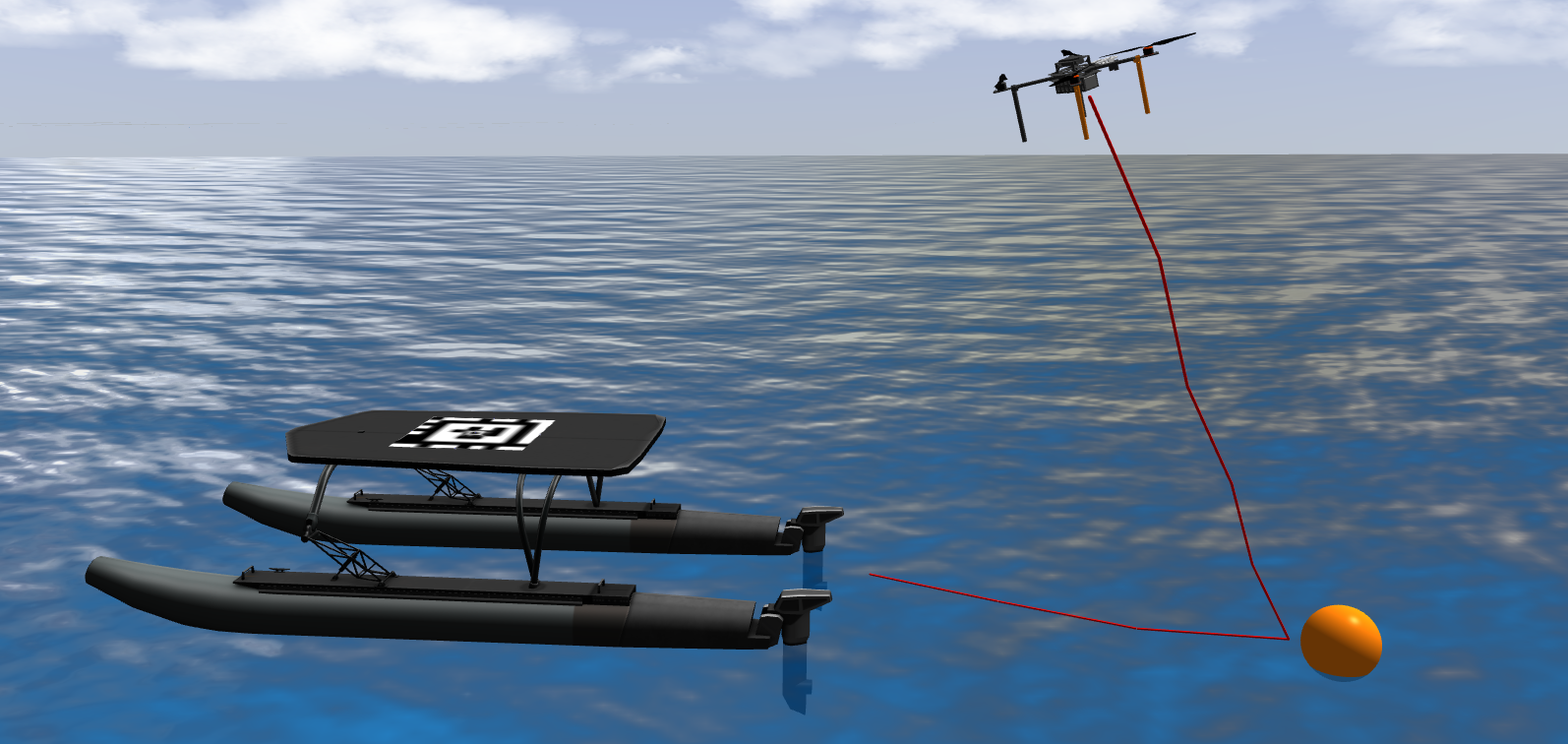}
    \caption{The multi-robot system consisting of a \ac{USV}, \ac{UAV}, and a solid object representing a sensor, garbage, or arbitrary floating object. This object is simultaneously tethered to the \ac{USV} and \ac{UAV}.}
    \label{fig:multi-robot_system}
\end{figure}

The robotic system presented in this paper consists of a solid object representing a sensor, garbage, or other arbitrary floating object connected via a tether to both a \ac{USV} and \ac{UAV} (see \reffig{fig:multi-robot_system}).
The main contributions of this paper are as follows:
\begin{itemize}
    \item We introduce a novel methodology for collaborative object manipulation on the surface of water by a team of \ac{UAV} and \ac{USV} using tethers. This is, according to our best knowledge, not currently presented in the available literature.
    \item  We present a novel mathematical model of the robotic system used in \ac{MPC} to precisely control the robots and move the object safely along the given path.
\end{itemize}
The presented multi-robot system is verified in numerous simulations in the realistic robotic simulator Gazebo and demonstrates its readiness to be deployed in the real world.

In the motivational scenario, we assume the presence of a compact \ac{USV} capable of towing an object across a water surface using a tether. 
The \ac{USV} carries a \ac{GNSS} receiver, \ac{IMU}, an onboard computer for robot movement control, and a mechanism for tether anchoring.
It is assumed that a compact \ac{UAV} is equipped with a \ac{GNSS} receiver, a flight controller featuring an \ac{IMU}, and an onboard computer for computing the desired commands for governing the \ac{UAV}-\ac{USV} team. 
The \ac{UAV} assumes the role of team commander, due to its situational awareness from an elevated perspective.
Additionally, we assume that the \ac{UAV} can estimate the position of the floating object using either onboard sensors or a communication link to the object, which itself may have the capability to estimate its own position.
Furthermore, a communication link between the \ac{UAV} and the \ac{USV} is essential in transferring the commands for the onboard controller from the \ac{UAV} to the \ac{USV} at a minimum frequency of 10~Hz.
The communication links within our framework may also be implemented as wired connections using the tethers that interconnect all components of our multi-robot system.

%%%%%%%%%%%%%%%%%%%%%%%%%%%%%%%%%%%%%%%%%%%%%%%%%%%%%%%%%%%%%%%%%%%%%%%%%%%%%%%%

\section{RELATED WORKS}
Collaborative object manipulation and transportation by multi-robot systems is a research topic in many scientific studies \cite{tuci2018CooperativeObjectTransport, bernard2011AutonomousTransportationDeployment, michael2011CooperativeManipulationTransportation, horyna2021AutonomousCollaborativeTransport}.
Autonomous transportation using a team of aerial robots is presented in \cite{bernard2011AutonomousTransportationDeployment, michael2011CooperativeManipulationTransportation, horyna2021AutonomousCollaborativeTransport}.
In addition to the above discussed payload and flight time limitations, as presented in \cite{bernard2011AutonomousTransportationDeployment}, transportation in strong winds causes stress on the robots, potentially leading to failure. 
Strong winds are often present above the water surface \cite{bagiorgas2012OffshoreWindSpeed}. 
Therefore, the usage of only aerial robots for object manipulation and transportation on water surfaces incurs a non-negligible risk of failure.
In our system, the \ac{USV} can offer a secure docking location in adverse weather conditions \cite{aissi2020AutonomousSolarUSV}.
Meanwhile, our framework allows the continuation of a given task using only the \ac{USV}, albeit at the expense of diminished system performance.

While the methods described in the previous paragraph involved lifting and carrying the object by the \acp{UAV} through the air, the approaches presented in \cite{brandao2022SidePullManeuverNovel, kourani2021BidirectionalManipulationBuoy, kourani2023ThreedimensionalModelingTethered} involve pulling the object along the surface using a rope, which is useful if the \ac{UAV} cannot lift the object.
However, the \ac{UAV} must exert significant force to pull the object, consuming a large amount of energy. 
Notably, this substantial energy consumption reduces the \ac{UAV} mission time, given the limited battery capacity of the \ac{UAV}.
Moreover, if the object is too heavy, the \ac{UAV} may not be able to pull it.
In our framework, the primary pulling force is provided by a \ac{USV}, which is typically more powerful than a \ac{UAV}.

The study of dynamics in a robotic system consisting of \ac{USV} and an \ac{AUV} connected by a cable is presented in \cite{vu2017StudyDynamicBehaviors, vu2020StudyDynamicBehavior}.
The \ac{AUV} is towed by the \ac{USV} for exploration purposes.
These studies show that a cable influences the motion of the robotics system, and that it is therefore necessary to model the mutual dynamics between individual components of the robotic system.
The work in \cite{chen2021DynamicPathPlanning} deals with path planning for a \ac{USV} towing a load in water using a rope, in which the concept of a safety boundary for towing is proposed to avoid obstacles.
This concept assumes that a towed object can be at any point near the \ac{USV} within a radius of the rope length.
However, this decreases the maneuverability of such a system and prolongs mission time, as the robot is required to maintain a greater distance from obstacles than strictly necessary.
Additionally, this concept also fails if the passage is narrower than the length of the rope.  
In our framework, the \ac{UAV} can quickly react to obstacles or narrow passages by adjusting the trajectory of the towed object to navigate around obstacles or pass through narrow passages safely.

A team of \ac{UAV} and \ac{USV} tethered together is introduced in \cite{talke2018CatenaryTetherShape}.
The study in \cite{talke2018CatenaryTetherShape} addresses the static catenary hanging cable problem and proposes a model for a catenary tether, aiming to enhance robustness and reduce power consumption in aerial robot flights. 
However, the study is mainly focused on cable modeling and the flying space of \ac{UAV}, and does not analyze the forces acting between the robots, tethers, and potential external floating objects.
Nevertheless, the power cable is useful for our proposed system as it solves one of the main limitations of the \acp{UAV} by providing energy from the \ac{USV} to the \ac{UAV}, significantly extending the \ac{UAV} mission time \cite{choi2014TetheredAerialRobots}.
Moreover, the cable enables the creation of a fast and reliable wired communication link between \ac{UAV} and \ac{USV} \cite{xu2021ApplicationResearchTethered}.

%%%%%%%%%%%%%%%%%%%%%%%%%%%%%%%%%%%%%%%%%%%%%%%%%%%%%%%%%%%%%%%%%%%%%%%%%%%%%%%%

\section{Dynamic system model}
\label{sec:system_model}
\begin{figure}[!t]
    \centering
    \includegraphics[width=\linewidth]{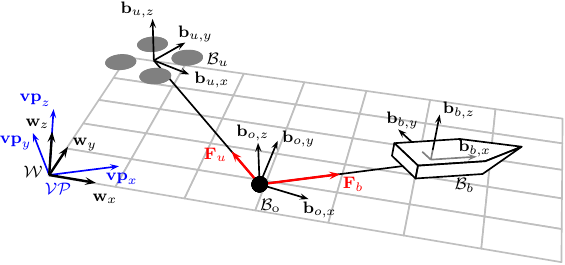}
    \caption{The depiction of the world frame $\worldframe = \{\worldframeaxis{x}, \worldframeaxis{y}, \worldframeaxis{z}\}$ in which the 3D position and orientation of the UAV $\bodyframe{u} = \{\bodyframeaxis{u}{x}, \bodyframeaxis{u}{y}, \bodyframeaxis{u}{z}\}$, USV $\bodyframe{b} = \{\bodyframeaxis{b}{x}, \bodyframeaxis{b}{y}, \bodyframeaxis{b}{z}\}$ and floating object $\bodyframe{o} = \{\bodyframeaxis{o}{x}, \bodyframeaxis{o}{y}, \bodyframeaxis{o}{z}\}$ body frames are expressed.  
    The red vectors $\uavforce$ and $\boatforce$ show the forces applied to the floating object by the UAV and USV tethers.
    The Vessel parallel coordinate system $\vesselparallelframe=\{ \vesselparallelframeaxis{x},\vesselparallelframeaxis{y},\vesselparallelframeaxis{z}\}$ is depicted in blue and is parallel to the USV body frame $\bodyframe{b}$.}
    \label{fig:system_sketch}
\end{figure}

This section presents a dynamic model of our multi-robot system consisting of the \ac{UAV}, \ac{USV}, and floating object.
The scheme of the system is shown in \reffig{fig:system_sketch}.
The $\worldframe = \{ \worldframeaxis{x}, \worldframeaxis{y}, \worldframeaxis{z} \}$ is the world frame common for all components of the multi-robot system, in which the position and rotation of the robots and the floating object are expressed.
The body frame of the floating object is denoted as $\bodyframe{o}$ with corresponding body-frame axes $\bodyframeaxis{o}{x},~\bodyframeaxis{o}{y},~\bodyframeaxis{o}{z}$.
Similarly, the $\bodyframe{u} = \{\bodyframeaxis{u}{x}, \bodyframeaxis{u}{y}, \bodyframeaxis{u}{z}\}$ is the body frame of the \ac{UAV} and $\bodyframe{b} = \{\bodyframeaxis{b}{x}, \bodyframeaxis{b}{y}, \bodyframeaxis{b}{z}\}$ is the body frame of the \ac{USV}.

\subsection{Floating object model}
As presented in \cite{Fossen2011}, the nonlinear equations describing the motion of a floating object on the water surface are
\begin{align}
    \vec{\dot{\eta}}_o &= \vec{J}_o(\vec{\eta}_o)\vec{\nu}_o,\\
    \vec{\tau}_o &=\vec{M}_o\vec{\dot{\nu}}_o + \vec{C}_o(\vec{\nu}_o)\vec{\nu}_o + \vec{D}_o(\vec{\nu}_o)\vec{\nu}_o + \vec{g}_o(\vec{\eta_o}), \label{eq:nonlinear_object_equation}
\end{align}
where $\vec{J}_o(\vec{\eta}_o)$ is the transformation from the body-fixed coordinate frame $\bodyframe{o}$ to the global coordinate frame $\mathcal{W}$, $\vec{M}_o = \vec{M}_{o,I}+\vec{M}_{o,A} \in \mathbb{R}^{6\times 6}$ is the sum of inertia matrix $\vec{M}_{o,I}$ and the hydrodynamic added mass matrix $\vec{M}_{o,A}$, presented due to the motion of the object through the fluid.
The $\vec{C}_o(\vec{\nu}_o) \in \mathbb{R}^{6\times 6}$ denotes the matrix of Coriolis and centripetal terms, $\vec{D}_o(\vec{\nu}_o) \in \mathbb{R}^{6\times 6}$ stands for the damping matrix, $\vec{g}_o(\vec{\eta}_o) \in \mathbb{R}^{6}$ denotes gravitational forces and torques, $\vec{\tau}_o \in \mathbb{R}^{6}$ represents the vector forces acting on the object by the \ac{USV} and \ac{UAV} through tethers.
The states of the object with respect to the coordinate frames presented above are
\begin{align}
    \vec{\eta}_o &= (\vec{p}^{\intercal}_o,\vec{\Theta}^{\intercal}_o)^{\intercal} = (x_o,y_o,z_o,\phi_o,\theta_o,\psi_o)^{\intercal},\label{eq:object_global_states}\\
    \vec{\nu}_o &= (\vec{v}^{\intercal}_o,\vec{\omega}^{\intercal}_o)^{\intercal} = (u_o,v_o,w_o,p_o,q_o,r_o)^{\intercal},\label{eq:object_body_states} 
\end{align}
where $\vec{\eta}_o$ stands for the position $\vec{p}_o=(x_o,y_o,z_o)^{\intercal}$ and orientation $\vec{\Theta}_o=(\phi_o,\theta_o,\psi_o)^{\intercal}$ in terms of the intrinsic Euler angles in a global coordinate frame $\worldframe$.
The vector $\vec{\nu}_o$ consists of linear velocity $\vec{v}_o=(u_o,v_o,w_o)^{\intercal}$ and angular velocity $\vec{\omega}_o=(p_o,q_o,r_o)^{\intercal}$ in a body-fixed coordinate frame $\bodyframe{o}$.

The following assumptions are made to simplify the model used in our controller:
\subsubsection{Low speed} In this paper, we assume that the object does not move at high speed. If the speed is low, the Coriolis and centripetal terms are neglected, and thus $\vec{C}_o(\vec{\nu}_o)\vec{\nu}_o=\vec{0}$.
\subsubsection{Negligible air drag} A floating object on the water has part of its body in the water and the rest above it. 
The drag of the water is much great than air drag.
Therefore, the damping term in \refeq{eq:nonlinear_object_equation} does not assume air drag.
\subsubsection{Linear damping, gravitational forces, and torques} The damping term and gravitational forces and torques in \refeq{eq:nonlinear_object_equation} are approximated by the linear terms \cite{Fossen2011}: $\vec{D}_o(\vec{\nu}_o)\vec{\nu}_o \approx \vec{D}_o\vec{\nu}_o$ and $\vec{g}_o(\vec{\eta}_o) \approx \vec{G}_o\vec{\eta}_o$.

Applying the presented assumptions, the \refeq{eq:nonlinear_object_equation} can be rewritten as
\begin{align}
    \vec{\dot{\nu}}_o &=\vec{M}_o^{-1} ( \vec{\tau}_o - \vec{D}_{o}\vec{\nu}_o - \vec{G}_{o}\vec{\eta}_o). \label{eq:linear_object_equation}
\end{align}

Two main forces are acting on the floating object through tethers: \ac{UAV} force $\vec{F}_{b}$ and \ac{USV} force $\vec{F}_{u}$ (see \reffig{fig:system_sketch}) as
\begin{align}
    \vec{\tau}_o &= \vec{F}_{u} + \vec{F}_{b},\\
    \vec{F}_{u} &= \vec{M}_{o} \vec{\dot{v}}_{u}^{\bodyframe{o}},\\
    \vec{F}_{b} &= \vec{M}_{o} \vec{\dot{v}}_{b}^{\bodyframe{o}},
\end{align}
where $\vec{\dot{v}}_{u}^{\bodyframe{o}} =(\dot{u}_u,\dot{v}_u,\dot{w}_u,0,0,0)^\intercal$ is the acceleration of the \ac{UAV} in $\bodyframe{o}$ frame, and $\vec{\dot{v}}_{b}^{\bodyframe{o}} = (\dot{u}_b,\dot{v}_b,0,0,0,0)^\intercal$ is the acceleration of the \ac{USV} in $\bodyframe{o}$ frame. 
Each force points in the direction of the corresponding stretched tether.
The tether is modeled as an inflexible, in-extensible connection between the floating object and the robot.
However, the complex tether model consisting of multiple segments can be integrated \cite{hong2020DynamicsModelingMotion}.
The following assumptions about the tether are considered for system modeling:
\subsubsection{Taut tether} The tether remains taut, i.e., $\norm{\vec{F}_u} > 0$ and $\norm{\vec{F}_b} > 0$.
Therefore, all components of the multi-robot system are coupled during the mission.
\subsubsection{No lifting} The object keeps floating on the water surface.
The vertical component of the force $\vec{\tau}_o$ does not cause the object to rise above the water surface, which is expressed as $\vec{\hat{w}}_{z}^\intercal\vec{J}_o(\vec{\eta_o})\vec{\tau}_o < m_o \norm{\vec{g}}$, where $m_o$ is the mass of the floating object, $\vec{g}$ is the gravitational acceleration, and $\vec{\hat{w}}_{z} = (0,0,1,0,0,0)^\intercal$.

%%%%%%%%%%%%%%%%%%%%%%%%%%%%%%%%%%%%%%%%%%%%%%%%%%%%%%%%%%%%%%%%%%%%%%%%%%%%%%%% START OF PIPELINE DIAGRAM: THIS IS THE FIGURE FOR CONTROL FRAMEWORK SECTION
\begin{figure*}[!t]
  \centering
  \resizebox{1.0\textwidth}{!}{
   \input{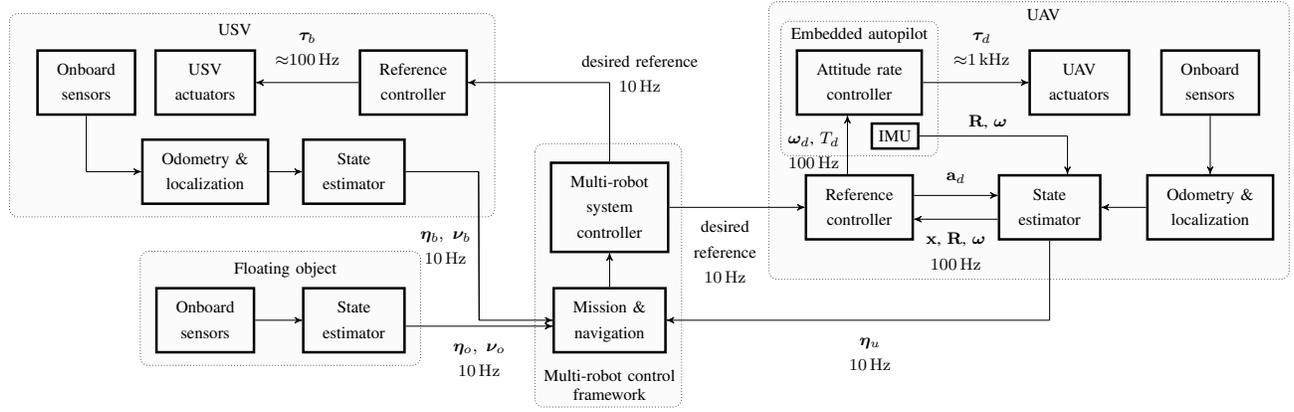}
   }
  \caption{
  Pipeline diagram of the proposed collaborative multi-robot control approach presented in this paper containing the MRS system \cite{baca2021MRSUAVSystem} for verification in realistic robotic scenarios. 
  \textit{Mission \& navigation} block decides about the reference of the floating object based on the current states and mission task.
  Then, \textit{Multi-robots system controller} computes the desired trajectory references using the MPC for the UAV and USV.
  The computed trajectory references are forwarded into the \textit{Reference controller} blocks of the UAV and USV.
  The UAV \textit{Reference controller} creates the desired thrust and angular velocities $(\bm{\omega}_d,T_d)$ for the Pixhawk embedded flight controller that commands the \textit{UAV actuators}.
  The UAV \textit{State estimator} fuses data from the \textit{Odometry \& localization} block to estimate the UAV translation, rotation, and angular velocities $(\mathbf{x},\mathbf{R},\bm{\omega})$.
  The USV \textit{Reference controller} creates the desired thrust $\vec{\tau}_b$ for \textit{USV actuators}.
  The USV \textit{State estimator} fuses data from the \textit{Odometry \& localization} block to estimate the USV states $(\vec{\eta}_b,\vec{\nu}_b)$.
  The floating object \textit{State estimator} fuses data from its \textit{Onboard sensors} to estimate states of the object $(\vec{\vec\eta}_o,\vec{\nu}_o)$.
  }
  \label{fig:pipeline_diagram}
\end{figure*}
%%%%%%%%%%%%%%%%%%%%%%%%%%%%%%%%%%%%%%%%%%%%%%%%%%%%%%%%%%%%%%%%%%%%%%%%%%%%%%%% END OF PIPELINE DIAGRAM:

\subsection{USV model}
The three \acp{DOF} \ac{USV} model considers surge $u_b$, i.e., motion in direction of $\bodyframeaxis{b}{x}$, sway $v_b$, i.e., motion in direction of $\bodyframeaxis{b}{y}$, and yaw angular rate $r_b$, which describes rotation around $\bodyframeaxis{b}{z}$. 
The states of the \ac{USV} model are
\begin{align}
    \vec{\eta}_b &= (x_b,y_b,\psi_b)^{\intercal},\label{eq:usv_global_states}\\
    \vec{\nu}_b &= (u_b,v_b,r_b)^{\intercal},\label{eq:usv_body_states}
\end{align}
where $x_b$ and $y_b$ stand for the position in the global coordinate frame $\mathcal{W}$, and yaw angle $\psi_b$ is defined in terms of the intrinsic Euler angle in $\mathcal{W}$.
The equations of \ac{USV} motion based on \cite{Fossen2011} are
\begin{align}
   \vec{\dot{\eta}}_b &= \vec{J}_b(\psi_b)\vec{\nu}_b,\\
   \vec{\dot{\nu}}_b &= (\vec{M}_{b,I}+\vec{M}_{b,A})^{-1}(\vec{B}_b\vec{u}_b - \vec{D}_b\vec{\nu}_b),
\end{align}
where $\vec{M}_{b,I}\in\mathbb{R}^{3\times 3}$ is the \ac{USV} inertia matrix, $\vec{M}_{b,A}\in\mathbb{R}^{3\times 3}$ denotes the hydrodynamic added mass, $\vec{D}_b\in\mathbb{R}^{3\times 3}$ is the \ac{USV} damping matrix, $\vec{u}_b = ( \tau_{\text{port}},\tau_{\text{starboard}})^\intercal$,
\begin{align}
    \vec{J}_b(\psi) &= \begin{pmatrix}
       \cos \psi_b & - \sin \psi_b & 0\\
       \sin \psi_b & \cos \psi_b & 0\\
       0 & 0 &  1
    \end{pmatrix},\\
    \vec{B}_b &= \begin{pmatrix}
        1 & 1\\
        0 & 0\\
        d_\tau/2 & -d_\tau/2
    \end{pmatrix}.
\end{align}
The $\tau_{\text{port}}$ represents the thrust generated by a motor located on the stern port side of the \ac{USV}, $\tau_{\text{starboard}}$ represents the thrust generated by a motor located on the stern starboard side of the \ac{USV}, and $d_\tau$ is a distance between the port and starboard motor.

\subsection{UAV model}
In order to create a \ac{UAV} model, it is assumed:
\subsubsection{Decoupled heading} The \ac{UAV} can move in any direction without changing its heading.
\subsubsection{Proper tether attachment} In our system, the tether is safely attached to the center of the \ac{UAV}.\\
These two assumptions allow us to consider only the position of the \ac{UAV} to model its dynamics.
If the second assumption does not hold (e.g., the tether is attached to one of the sides of the \ac{UAV}), it becomes necessary to model the force exerted when the \ac{UAV} changes its heading, even if it maintains its position.
The state of the \ac{UAV} in the $\worldframe$ frame is expressed as
\begin{align}
    \vec{\eta}_u = (x_u,u_u,y_u,v_u,z_u,w_u)^\intercal,
\end{align}
where $\vec{p}_u=(x_u,y_u,z_u)^\intercal$ is the \ac{UAV} position and $\vec{v}_u=(u_u,v_u,w_u)^\intercal$ is the \ac{UAV} velocity.
Similarly to \cite{baca2021MRSUAVSystem}, the \ac{UAV} dynamics are modeled as a point-mass double integrator expressed as a decoupled \ac{LTI} system
\begin{align}
    \vec{\dot{\eta}}_u &= \vec{A}_u \vec{\eta}_u + \vec{B}_u \vec{u}_u,
\end{align}
where the term $\vec{u}_u=(u_{u,u}, u_{u,v}, u_{u,w})$ is an input vector for the \ac{UAV} model, and
\begin{align}
    \vec{A}_u = \begin{pmatrix}
        \vec{A}_1 & \vec{0}^{2\times2}   & \vec{0}^{2\times2}\\
        \vec{0}^{2\times2}   & \vec{A}_1 & \vec{0}^{2\times2}\\
        \vec{0}^{2\times2}   & \vec{0}^{2\times2}   & \vec{A}_1\\
    \end{pmatrix},~
    \vec{B}_u
    \begin{pmatrix}
        \vec{B}_1 & \vec{0}^{2\times1}& \vec{0}^{2\times1}\\
        \vec{0}^{2\times1} & \vec{B}_1 & \vec{0}^{2\times1}\\
        \vec{0}^{2\times1} & \vec{0}^{2\times1} & \vec{B}_1
    \end{pmatrix}.
\end{align}
Subsystem matrices $\vec{A}_1$ and $\vec{B}_1$ are denoted as
\begin{align}
     \vec{A}_1 = \begin{pmatrix}
        0 & a_1\\
        0 & 0\\
    \end{pmatrix},~\vec{B}_1 = \begin{pmatrix}
        0\\
        b_1\\
    \end{pmatrix},
\end{align}
where $a_1$ and $b_1$ are parameters for the subsystem of the \ac{UAV} model.

%%%%%%%%%%%%%%%%%%%%%%%%%%%%%%%%%%%%%%%%%%%%%%%%%%%%%%%%%%%%%%%%%%%%%%%%%%%%%%%%
\section{Control framework of the system}

The control framework of the system aims to safely guide the floating object on the water surface alongside the given reference and be resistant to external disturbances, such as water currents and winds.
The pipeline diagram of the proposed control framework is shown in \reffig{fig:pipeline_diagram}.
The diagram contains four main areas that represent the floating object, \ac{USV}, \ac{UAV}, and the multi-robot control framework.

The mission and navigation block takes the current states of all multi-robot system components $\vec{\vec\eta}_o$, $\vec{\nu}_o$, $\vec{\vec\eta}_b$, $\vec{\nu}_b$, $\vec{\vec\eta}_u$ from the corresponding state estimators using data from onboard sensors.
However, the onboard sensors of the floating object can be easily replaced by detection methods using images from a camera placed on the \ac{UAV} or \ac{USV} \cite{stepan2019jfr} if onboard sensor data are not available.
Further, the mission and navigation block decides about the next floating object reference according to the ongoing mission.
The multi-robot system controller then initializes the linear \ac{MPC} solver \cite{osqp} by the states $\vec{\vec\eta}_o$, $\vec{\nu}_o$, $\vec{\vec\eta}_b$, $\vec{\nu}_b$, $\vec{\vec\eta}_u$ and the model of the system presented in \refsec{sec:system_model}.
The \ac{MPC} commands the \ac{UAV} and \ac{USV} to move the floating object alongside the desired trajectory reference.

The \ac{MPC} needs a linear discrete model of the system; therefore, the \ac{RK4} can be used to acquire such a model from the one presented in \refsec{sec:system_model}.
However, our application allows us to make several assumptions to simplify the model of the object.
Firstly, the roll $\phi_o$ and pitch $\theta_o$ angles are assumed to be small, i.e., it is not necessary to consider it in the \ac{MPC}.
In our case, we can also assume an almost symmetrical rigid body around the $\bodyframeaxis{o}{z}$ axis. 
Moreover, the floating object is connected to a tether under tension, which causes $\psi_o$ stabilization by the system dynamics itself. 
Therefore, we can neglect the $\psi_o$ as well, and then only the position $\vec{p}_o$ and the velocity $\vec{v}_o^{\worldframe}$ of the floating object in the $\worldframe$ frame is considered in the \ac{MPC} without needs to use $\vec{J}_o(\vec{\eta}_o)$ transformation.
Lastly, the \textit{Vessel parallel coordinate system} $\vesselparallelframe$ is defined as $\vec{\eta}_b^{\vesselparallelframe} = \vec{J}_b^\intercal(\psi_b)\vec{\eta}_b$, which results to $\vec{\dot{\eta}}_b^{\vesselparallelframe}\approx\vec{\nu}_b$ (see \reffig{fig:system_sketch}).
Therefore, all vectors in our control framework are transformed into $\vesselparallelframe$. Then, our dynamic system model is linear, as the linear \ac{MPC} requires. 
The final model of the proposed multi-robot system, detailed in \refsec{sec:system_model}, and used in the \ac{MPC}, is
\begin{align}
    \vec{\dot{p}}_o^{\vesselparallelframe} &= \vec{v}_o^{\vesselparallelframe},\label{eq:first_overall_model}\\
    \vec{\dot{v}}_o^{\vesselparallelframe} &=\vec{M}_o^{-1} (- \vec{D}_{o}\vec{v}_o^{\vesselparallelframe} - \vec{G}_{o}\vec{p}_o^{\vesselparallelframe} + \vec{M}_o(\vec{\dot{v}}_u^{\vesselparallelframe} + \vec{\dot{v}}_b^{\vesselparallelframe})),\\
    \vec{\dot{\eta}}_b^{\vesselparallelframe} &= \vec{\nu}_b,\\
   \vec{\dot{\nu}}_b &= (\vec{M}_{b,I}+\vec{M}_{b,A})^{-1}(\vec{B}_b\vec{u}_b - \vec{D}_b\vec{\nu}_b),\\
   \vec{\dot{\eta}}_u^{\vesselparallelframe} &= \vec{A}_u \vec{\eta}_u^{\vesselparallelframe} + \vec{B}_u \vec{u}_u^{\vesselparallelframe},\label{eq:last_overall_model}
\end{align}
where $\vec{v}_b^{\vesselparallelframe} = (u_b,v_b,0)^\intercal$ is a \ac{USV} planar velocity in $\vesselparallelframe$.

The linear continuous model \refeq{eq:first_overall_model}--\refeq{eq:last_overall_model} is discretized as 
\begin{align}
    \vec{x}_{[k+1]} = \vec{A}\vec{x}_{[k]} + \vec{B}\vec{u}_{[k]},
\end{align}
where $\vec{x}_{[k]}^\intercal = (\vec{p}_o^\intercal, \vec{v}_o^\intercal, \vec{\eta}_b^\intercal, \vec{\nu}_b^\intercal, \vec{\eta}_u^\intercal)^{\vesselparallelframe}_{[k]}$ is the state of the model of the multi-robot system expressed in $\vesselparallelframe$ at time step $k$, and $\vec{u}_{[k]} = (\vec{u}_b^\intercal,\vec{u}_u^\intercal)^{\vesselparallelframe}_{[k]}$ is the input of the model expressed in $\vesselparallelframe$ at time step $k$.
The control error for the prediction horizon $n\in\mathbb{Z}^{+}$ is defined as
\begin{align}
    \vec{e}_{[k]} = \vec{x}_{[k]} - \vec{x}_{r,[k]} \quad \forall k\in\{1,...,n\},
\end{align}
where $\vec{x}_{r,[k]}$ is the reference for the state $\vec{x}_{[k]}$ at time step $k$.
The \ac{MPC} is defined as a quadratic programming problem
\begin{align}
    \min_{\vec{u}_{[1:n]}} \sum_{k=1}^{n-1}\left( \vec{e}^{\intercal}_{[k]}\vec{Q}\vec{e}_{[k]} + \vec{u}^{\intercal}_{[k]}\vec{R}\vec{u}_{[k]} \right) + \vec{e}^{\intercal}_{[n]}\vec{S}\vec{e}_{[n]}
\end{align}
\begin{align}
    \text{s.t.}\quad
    \vec{x}_{[k+1]} = \vec{A}\vec{x}_{[k]} + \vec{B}\vec{u}_{[k]} \qquad &\forall k\in\{1,...,n\},\\
    \vec{x}_{\min} \leq \vec{x}_{[k]} \leq \vec{x}_{\max}\qquad&\forall k\in\{1,...,n\},\\
    \vec{u}_{\min} \leq \vec{u}_{[k]} \leq \vec{u}_{\max}\qquad&\forall k\in\{1,...,n-1\},
\end{align}
where $\vec{x}_{\min}$ is the lower bound for state $\vec{x}_{[k]}$, $\vec{x}_{\max}$ is the upper bound for state $\vec{x}_{[k]}$, $\vec{u}_{\min}$ is the lower bound for input $\vec{u}_{[k]}$, and $\vec{u}_{\max}$ is the upper bound for input $\vec{u}_{[k]}$.
The penalization matrices $\vec{Q}$, $\vec{R}$, and $\vec{S}$ are positive semi-definite.
% The \ac{MPC} also limits the state values of \ac{UAV}, \ac{USV}, and floating object to reach the constraints of the real robots.
Additional constraints are also applied to the \ac{UAV} motion to keep the tether under tension, making the object motion fully controllable.
The \ac{UAV} motion at fixed altitude is limited to a circle, with the floating object at the center position and a radius equal to tether length $l_t$, relaxed by small constant $\epsilon$.
This results in two-circle boundaries with radii $l_t+\epsilon$ and $l_t-\epsilon$ that are linearized and added to the \ac{MPC} constraints:
\begin{align}
   (-y_u,x_u)\vec{v}_{cb} = (-p_{\min,y},p_{\min,x})\vec{v}_{cb},\label{eq:boundary_min}\\ 
   (-y_u,x_u)\vec{v}_{cb} = (-p_{\max,y},p_{\max,x})\vec{v}_{cb},\label{eq:boundary_max} 
\end{align}
where the vector $\vec{v}_{cb}$ is perpendicular to the vector $\vec{u}_{cb}^\intercal=(x_u,y_u)-(x_o,y_o)$.
The values $p_{\min,x}$,~$p_{\min,y}$,~$p_{\max,x}$, and $p_{\max,y}$ are computed as
\begin{align}
    (p_{\min,x},p_{\min,y}) = \left(\sqrt{l_t^2-(z_u-z_o)^2}-\epsilon\right)\dfrac{\vec{u}_{cb}^\intercal}{\norm{\vec{u}_{cb}}},\\
    (p_{\max,x},p_{\max,y}) = \left(\sqrt{l_t^2-(z_u-z_o)^2}+\epsilon\right)\dfrac{\vec{u}_{cb}^\intercal}{\norm{\vec{u}_{cb}}}.
\end{align}
The \ac{UAV} position $(x_u,y_u)$ has to lie between these two linear boundaries \refeq{eq:boundary_min}--\refeq{eq:boundary_max}.
If these constraints specified by \refeq{eq:boundary_min}--\refeq{eq:boundary_max} are violated, the $\epsilon$ constant can be enlarged.

The Multi-robot system controller outputs the desired trajectories for the \ac{UAV} and \ac{USV} sent to their Reference controllers at 10~Hz (see \reffig{fig:pipeline_diagram}). 
The \ac{USV} reference controller based on \cite{Fossen2011} runs at 100~Hz and outputs two values: thrusts $\vec{\tau}_{\text{port}}$ and $\vec{\tau}_{\text{starboard}}$ for the corresponding motors located on the \ac{USV} stern pointing to the opposite direction of the $\bodyframeaxis{b}{x}$ axis.
The \ac{UAV} controller is made up of the \ac{MRS} \ac{UAV} system \cite{baca2021MRSUAVSystem}, which uses a complex dynamic \ac{UAV} model, therefore, enabling precise trajectory tracking.

%%%%%%%%%%%%%%%%%%%%%%%%%%%%%%%%%%%%%%%%%%%%%%%%%%%%%%%%%%%%%%%%%%%%%%%%%%%%%%%%

\section{Results}
\begin{figure}[!b]
  \centering
    \includegraphics[width=\linewidth]{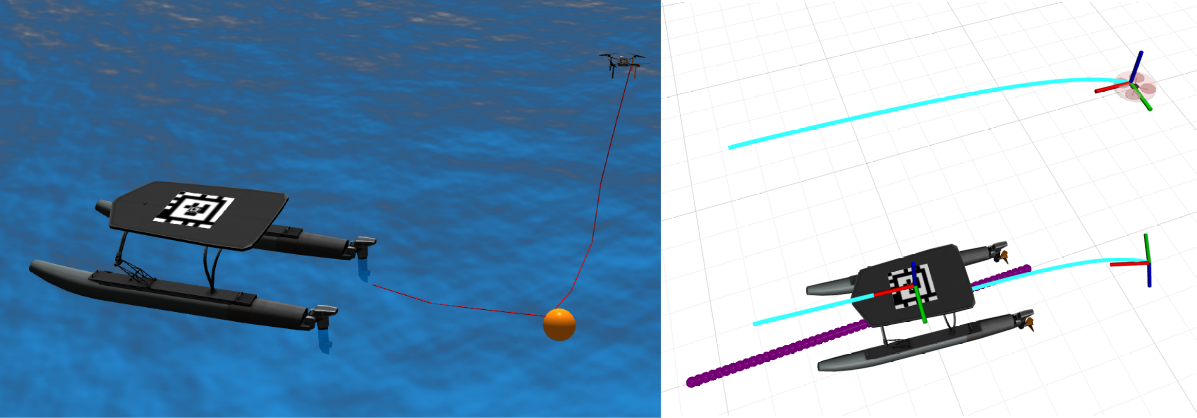}
  \caption{Our multi-robot system, comprising the UAV, USV, and floating object, is depicted in the Gazebo simulator and RViz during its mission.}
  \label{fig:gazebo_and_rviz}
\end{figure}

The presented approach was extensively tested in the realistic robotic simulator Gazebo, and extended by the \ac{VRX} simulator to imitate a marine environment \cite{bingham19toward} (see \reffig{fig:gazebo_and_rviz}).
From all experiments, we have selected two situations - trajectory tracking and response to a disturbance, which have best illustrated the differences between our multi-robot approach and the state-of-the-art single-robot approach based on \cite{vu2020StudyDynamicBehavior}.
Overall, the results of numerous experiments show us that the multi-robot system is prepared for real-world deployment.
The multimedia materials supporting the results of this paper are available at \url{https://mrs.felk.cvut.cz/papers/uav-usv-object-manipulation}.

\subsection{Experimental setup}
The \ac{WAM-V} platform \cite{bingham19toward} has been chosen as the \ac{USV} and is linked to the object by a tether with a length of 4~m.
The object is modeled as a sphere with a radius of 0.25~m and a mass of 5~kg.
The Tarot T650 platform is used as the \ac{UAV} \cite{HertJINTHW_paper} and is tethered to the object with a 5~m tether.

\subsection{Trajectory tracking}
\label{sec:trajectory_tracking}
\begin{figure}[!t]
  \centering
    \includegraphics[width=\linewidth]{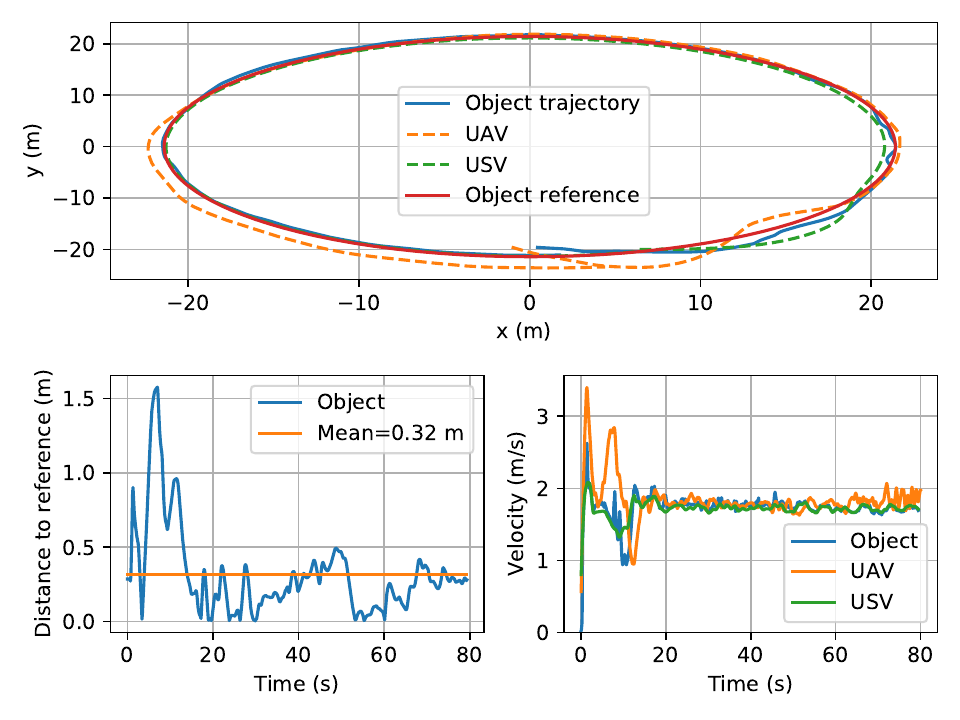}
  \caption{The results of the experiment, where our multi-robot system tracks a circular trajectory for the object.
  The graphs illustrate the position of the multi-robot system and the reference in a horizontal plane, as well as the distance of the object to the reference and the velocities of the UAV, USV, and the object.}
  \label{fig:exp_circle}
\end{figure}

The trajectory for the robots to follow can be effectively created using only straight line segments and circular arcs \cite{dubins}.
Therefore, we evaluate our system on the trajectories consisting of these two types.
Evaluation of straight-line trajectories shows that our multi-robot system and the single-robot approach \cite{vu2020StudyDynamicBehavior} have identical performance in case of no disturbances.
The average distance of the object to the trajectory reference for our multi-robot system is 0.08~m, and it is 0.09 m for the single-robot approach.
Therefore, we moved to track a circular reference trajectory for the object, as it clearly showcases the difference between the single-robot approach \cite{vu2020StudyDynamicBehavior} and our multi-robot approach. 
The experiment is illustrated in \reffig{fig:exp_circle}.
The reference trajectory initiates from the position (0,-20) and proceeds in a counterclockwise direction.
Initially, the robots must approach the trajectory.
This approaching phase showcases the behavior of the \ac{UAV} flying to the opposite side of the reference trajectory compared to the \ac{USV} in order to keep the object close to the reference.
Subsequently, the object is securely guided along the reference trajectory, with the system operating smoothly at a constant speed.
The average distance of the object to the reference is 0.32~m, with the largest distance occurring at the beginning when the system is approaching the reference.

\begin{figure}[!t]
  \centering
    \includegraphics[width=\linewidth]{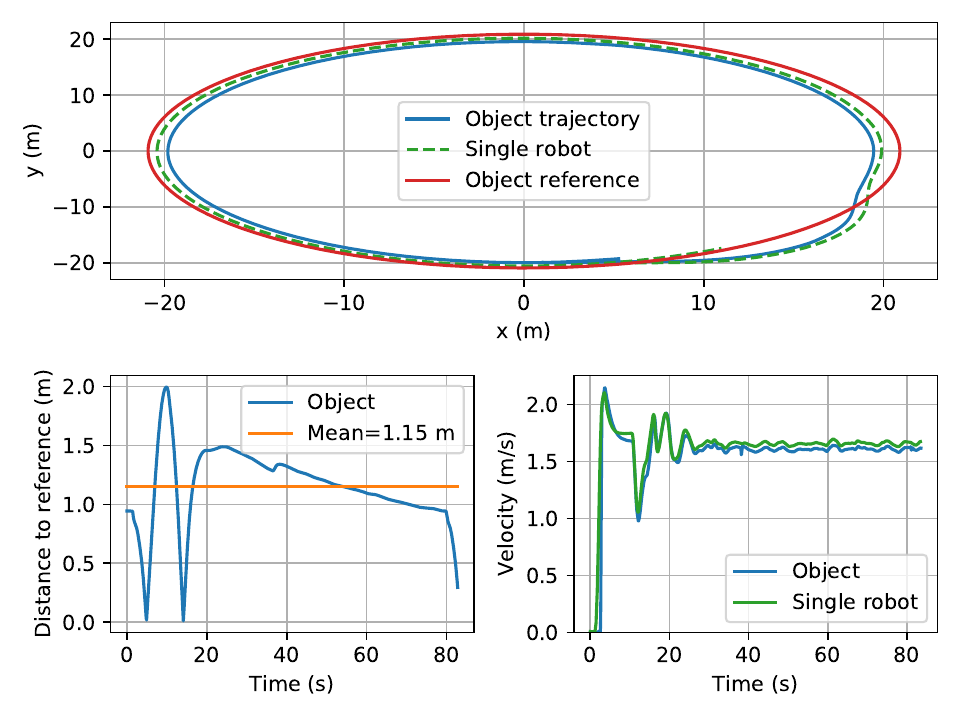}
  \caption{The results of the experiment, where a single robot tracks a circular trajectory for the object.
  The graphs depict the position of the single robot, the object, and the reference in a horizontal plane, as well as the distance of the object to the reference and the velocities of the single robot and the object.
  }
  \label{fig:exp_circle_usv_only}
\end{figure}

To compare our multi-robot system with the single-robot approach \cite{vu2020StudyDynamicBehavior}, the same trajectory is tracked using only a single robot (the \ac{USV} in our case) as we assume that the object is too heavy for a single \ac{UAV}.
The results are shown in \reffig{fig:exp_circle_usv_only}.
The mean distance of the object from the reference is 1.15~m, which is 3.5 times larger than in our multi-robot system.
Additionally, as illustrated in \reffig{fig:exp_circle_usv_only}, the approach to the reference takes twice as long in the case of a single robot.
The results show that our multi-robot system is more efficient than a single robot, as the \ac{UAV} acting on the object together with the \ac{USV} reduces the reference tracking error. 

\subsection{Disturbance response}
In the upcoming experiments, we demonstrate how both our multi-robot system and single-robot approach \cite{vu2020StudyDynamicBehavior} respond to external disturbances. The disturbance is applied as a force acting on the object perpendicular to its movement. Various sources of disturbance on the water surface, such as currents, winds, or even animals, can contribute to this force. The disturbance force is applied to the object for a duration of 0.5~s with a magnitude of 1~kN.

\begin{figure}[!t]
  \centering
    \includegraphics[width=\linewidth]{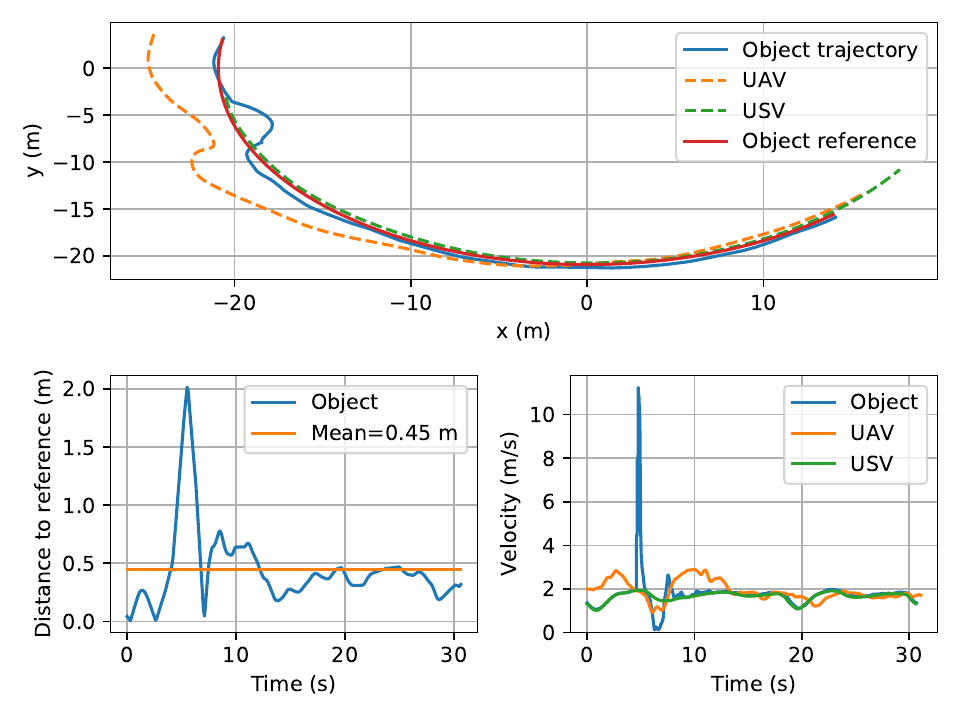}
  \caption{The results of the experiment for the multi-robot system, in which disturbance force affects the object with a magnitude 1~kN for 0.5~s. 
  The graphs depict the position of the multi-robot system and the reference in a horizontal plane, as well as the distance of the object to the reference and the velocities of the UAV, USV, and the object.
  }
  \label{fig:exp_disturbance}
\end{figure}

The \reffig{fig:exp_disturbance} illustrates how our multi-robot system responds to the disturbance.
The moment of acting disturbance force is clearly visible in the graph of velocities as a large peak in object velocity.
The object is then displaced from its trajectory.
However, the multi-robot system recovers from the disturbance after 3.3~s.
Thereafter, the multi-robot system smoothly follows the trajectory reference for the object.

\begin{figure}[!t]
  \centering
    \includegraphics[width=\linewidth]{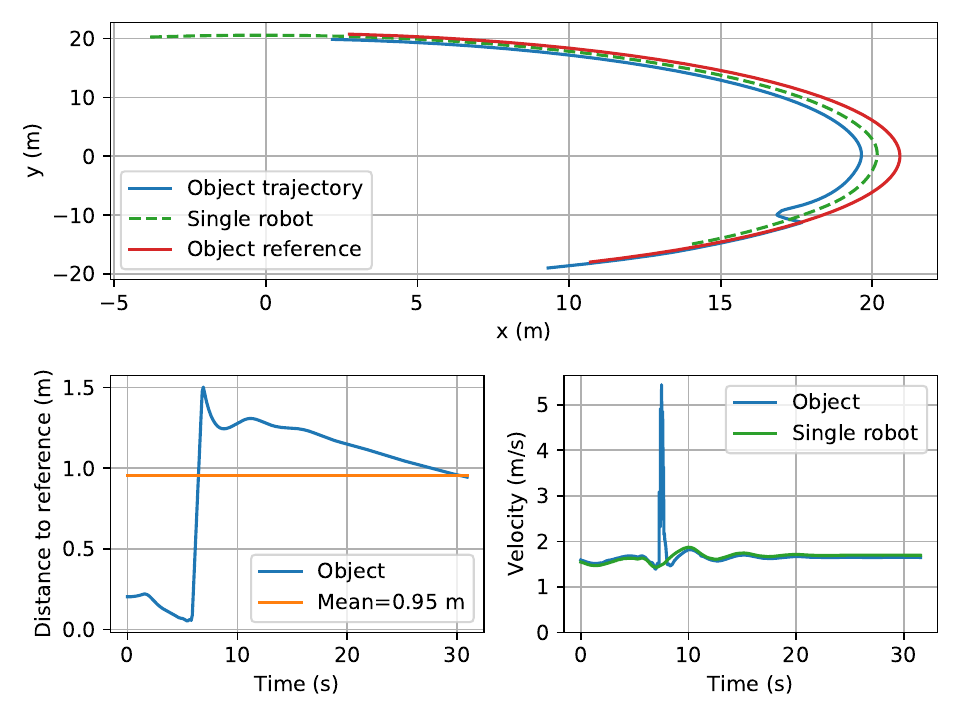}
  \caption{The results of the experiment for the single-robot approach, in which disturbance force affects the object with a magnitude 1~kN for 0.5~s. 
  The graphs depict the position of the single robot, the object, and the reference in a horizontal plane, as well as the distance of the object to the reference and the velocities of the single robot and the object.
  }
  \label{fig:exp_disturbance_usv_only}
\end{figure}

The same disturbance has been applied to the single-robot system, as shown in \reffig{fig:exp_disturbance_usv_only}.
The disturbance occurs at time 7~s, and the single-robot system again reaches the mean distance 1.15~m taken from the trajectory tracking experiment (\reffig{fig:exp_circle_usv_only}) at time 20~s.
Therefore, the recovery time, equal to 13~s, is four times longer compared to the multi-robot system.
Moreover, the mean distance of the object from the reference is twice as large, reflecting the increased difficulty in realigning the sensor with the reference trajectory for a single robot.
 
In the conducted experiments, both approaches are required to perform aggressive maneuvers for fast recovery from the disturbance, such as turning on a small turning radius. 
During these aggressive maneuvers, our multi-robot system outperforms the single-robot approach, as our system has both a smaller distance from object to reference and a shorter recovery time from disturbance.
The better performance of our system is attributed to the collaborative efforts of the \ac{USV} pulling the object forward towards the reference, while the \ac{UAV} adjusts the object position laterally to achieve closer alignment with the reference trajectory.

\subsection{Statistical evaluation}
\begin{figure}[!t]
  \centering
    \includegraphics[width=0.8\linewidth]{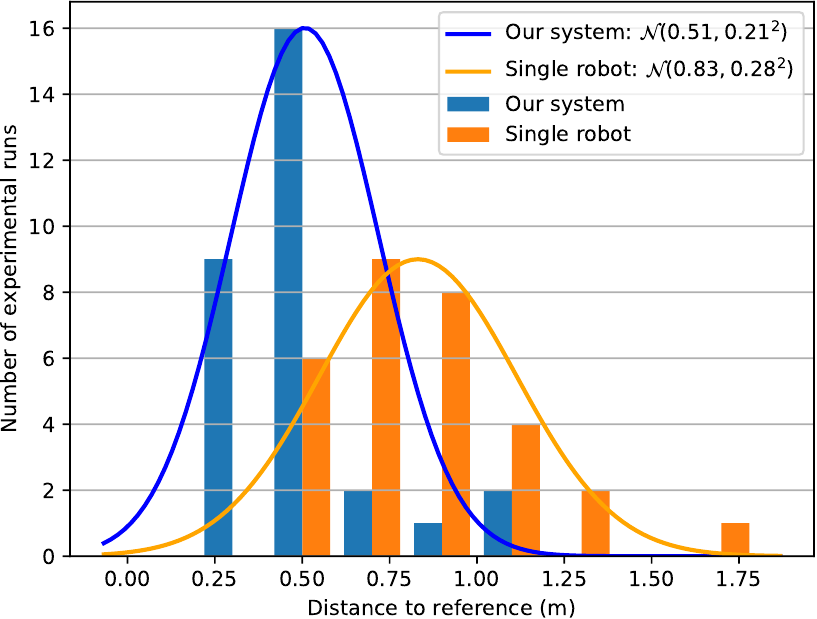}
  \caption{Statistical evaluation of the proposed multi-robot system and the single-robot approach. The graph shows the mean distances of the object to the trajectory reference across individual experimental runs for each method, subsequently fitted to the normal distribution.
  }
  \label{fig:histogram}
\end{figure}

This section presents a statistical evaluation of the proposed \systemname.
We conducted over 60 experiments on trajectories comprising both straight line segments and circular arcs, the use of which is explained in the previous section \ref{sec:trajectory_tracking}.
The mean distance of the object to the reference trajectory for both approaches across individual experimental runs is proposed in \reffig{fig:histogram}.
Subsequently, the distance values are fitted to the normal distribution, which provides us with the mean value and standard deviation for each method.

The mean value of the normal distribution for our multi-robot system is 0.51~m, noticeably smaller than the mean value associated with the single-robot approach \cite{vu2020StudyDynamicBehavior} (0.83~m). 
Therefore, our multi-robot approach maintains smaller tracking errors during the conducted experiments.
Furthermore, the standard deviation of our multi-robot system (0.21~m) is found to be less than that of the single-robot approach (0.28~m).
In summary, our new multi-robot system demonstrates better performance than the single-robot approach by producing smaller trajectory tracking errors with reduced variability across different trajectory references.

The \reffig{fig:histogram} also reveals that in two experimental runs of our multi-robot system, the mean distance to the trajectory reference exceeded 1~m.
Our observations indicate that this occurs when the \ac{UAV} is positioned on one side of the tether, but needs to apply force to the other side.
Consequently, the \ac{UAV} must first relocate to the appropriate side to tension the tether in the correct direction before influencing the object's movement with force.
The transition of the \ac{UAV} from one side to the other takes some time, during which the object is not aligned with its reference.
This temporal misalignment results in a larger mean distance from the trajectory reference during this interval and occurs because of the condition that the tethers must remain taut at all times otherwise, it would be faster.

%%%%%%%%%%%%%%%%%%%%%%%%%%%%%%%%%%%%%%%%%%%%%%%%%%%%%%%%%%%%%%%%%%%%%%%%%%%%%%%%

\section{Conclusion}
We have introduced a novel control framework designed for collaborative object manipulation on the surface of water using a \ac{UAV}-\ac{USV} team.
The complex dynamics of the multi-robot system, formed by the \ac{UAV} and \ac{USV} tethered to the object, are studied in this paper.
Based on this analysis, we propose a novel mathematical model for the multi-robot team tethered to the object.
The model is used in an \ac{MPC} fashion to showcase the ability of our framework to manipulate the object on the water surface by the \ac{UAV}-\ac{USV} team.
Numerous experiments were conducted to demonstrate our multi-robot system's performance and show that the system is prepared for real-world deployment.
The comparative analysis between our multi-robot system and a single-robot approach reveals that our multi-robot system exhibits a \SI{40}{\percent} lower tracking error and a four times shorter recovery time from disturbances. 
These findings align with our hypothesis, affirming that a multi-robot team is more efficient for this task.

%%%%%%%%%%%%%%%%%%%%%%%%%%%%%%%%%%%%%%%%%%%%%%%%%%%%%%%%%%%%%%%%%%%%%%%%%%%%%%%%

\bibliographystyle{IEEEtran}
\bibliography{main}

%%%%%%%%%%%%%%%%%%%%%%%%%%%%%%%%%%%%%%%%%%%%%%%%%%%%%%%%%%%%%%%%%%%%%%%%%%%%%%%%

\begin{acronym}
  \acro{CNN}[CNN]{Convolutional Neural Network}
  \acro{CPU}[CPU]{Central Processing Unit}
  \acro{IR}[IR]{infrared}
  \acro{GNSS}[GNSS]{Global Navigation Satellite System}
  \acro{MOCAP}[mo-cap]{Motion capture}
  \acro{MPC}[MPC]{Model Predictive Control}
  \acro{MRS}[MRS]{Multi-robot Systems group}
  \acro{ML}[ML]{Machine Learning}
  \acro{MAV}[MAV]{Micro-scale Unmanned Aerial Vehicle}
  \acro{UAV}[UAV]{Unmanned Aerial Vehicle}
  \acro{UV}[UV]{ultraviolet}
  \acro{UVDAR}[\emph{UVDAR}]{UltraViolet Direction And Ranging}
  \acro{UT}[UT]{Unscented Transform}
  \acro{RTK}[RTK]{Real-Time Kinematic}
  \acro{ROS}[ROS]{Robot Operating System}
  \acro{wrt}[w.r.t.]{with respect to}
  \acro{LTI}[LTI]{Linear time-invariant}
  \acro{USV}[USV]{Unmanned Surface Vehicle}
  \acroplural{DOF}[DOFs]{Degrees of Freedom}
  \acro{DOF}[DOF]{Degree of Freedom}
  \acro{API}[API]{Application Programming Interface}
  \acro{CTU}[CTU]{Czech Technical University}
  \acroplural{DOF}[DOFs]{Degrees of Freedom}
  \acro{DOF}[DOF]{Degree of Freedom}
  \acro{FOV}[FOV]{Field of View}
  \acro{GNSS}[GNSS]{Global Navigation Satellite System}
  \acro{GPS}[GPS]{Global Positioning System}
  \acro{IMU}[IMU]{Inertial Measurement Unit}
  \acro{LKF}[LKF]{Linear Kalman Filter}
  \acro{LTI}[LTI]{Linear time-invariant}
  \acro{LiDAR}[LiDAR]{Light Detection and Ranging}
  \acro{MAV}[MAV]{Micro Aerial Vehicle}
  \acro{MPC}[MPC]{Model Predictive Control}
  \acro{MRS}[MRS]{Multi-robot Systems}
  \acro{ROS}[ROS]{Robot Operating System}
  \acro{RTK}[RTK]{Real-time Kinematics}
  \acro{SLAM}[SLAM]{Simultaneous Localization And Mapping}
  \acro{UAV}[UAV]{Unmanned Aerial Vehicle}
  \acro{UGV}[UGV]{Unmanned Ground Vehicle}
  \acro{UKF}[UKF]{Unscented Kalman Filter}
  \acro{USV}[USV]{Unmanned Surface Vehicle}
  \acro{RMSE}[RMSE]{Root Mean Square Error}
  \acro{UVDAR}[UVDAR]{UltraViolet Direction And Ranging}
  \acro{UV}[UV]{UltraViolet}
  \acro{VRX}[VRX]{Virtual RobotX}
  \acro{WAM-V}[WAM-V]{Wave Adaptive Modular Vessel}
  \acro{GT}[GT]{Ground Truth}
  \acro{UTM}[UTM]{Universal Transverse Mercator}
  \acro{AUV}[AUV]{Autonomous Underwater Vehicle}
  \acro{RK4}[RK4]{Runge--Kutta method of fourth order}
\end{acronym}

\end{document}